\documentclass[sigconf, review=false, anonymous=false]{acmart}

\copyrightyear{2026}
\acmYear{2026}
\setcopyright{rightsretained}

\acmConference[SIGIR '26]{Proceedings of the 49th International ACM SIGIR Conference on Research and Development in Information Retrieval}{July 20--24, 2026}{Melbourne, Australia.}
\acmBooktitle{Proceedings of the 49th Int'l ACM SIGIR Conference on Research and Development in Information Retrieval (SIGIR '26), July 20--24, 2026, Melbourne, Australia}

\makeatother

\usepackage{enumitem}
\newlist{inlinelist}{enumerate*}{1}
\setlist*[inlinelist,1]{%
  label=(\roman*),
}
\usepackage{subfigure}
\usepackage{adjustbox}
\usepackage{booktabs}
\usepackage{multicol}
\usepackage{multirow}
\usepackage{xspace}
\usepackage{algorithm}
\usepackage{algpseudocode}

\newcommand{\ourmethod}{\texttt{$P^3$}\xspace}

\title{Improving User Privacy in Personalized Generation}
\subtitle{Client-Side Retrieval-Augmented Modification of Server-Side Generated Speculations}

\author{Alireza Salemi}
\affiliation{%
  \institution{University of Massachusetts Amherst}
  \city{Amherst}
  \state{MA}
  \country{USA}
}
\email{asalemi@cs.umass.edu}

\author{Hamed Zamani}
\affiliation{%
  \institution{University of Massachusetts Amherst}
  \city{Amherst}
  \state{MA}
  \country{USA}
}
\email{zamani@cs.umass.edu}

\begin{document}


\begin{abstract}
Personalization is crucial for aligning Large Language Model (LLM) outputs with individual user preferences and background knowledge. State-of-the-art solutions are based on retrieval augmentation where relevant context from user profile is retrieved for LLM consumption. These methods deal with a trade-off between exposing retrieved private data to cloud providers and relying on less capable local models. We introduce \ourmethod{}, an interactive framework for high-quality personalization without revealing private profiles to server-side LLMs. In \ourmethod{}, a large server-side model  generates a sequence of $k$ draft tokens based solely on the user query, while a small client-side model, with retrieval access to the user's private profile, evaluates and modifies these drafts to better reflect user preferences. This process repeats until an end token is generated. Experiments on LaMP-QA, a recent benchmark consisting of three personalized question answering datasets, show that \ourmethod{} consistently outperforms both non-personalized server-side and personalized client-side baselines, achieving statistically significant improvements of $7.4\%$ to $9\%$ on average. Importantly, \ourmethod{} recovers $90.3\%$ to $95.7\%$ of the utility of a ``leaky'' upper-bound scenario in which the full profile is exposed to the large server-side model. Privacy analyses, including linkability and attribute inference attacks, indicate that \ourmethod{} preserves the privacy of a non-personalized server-side model, introducing only marginal additional leakage ($1.5\%$–$3.5\%$) compared to submitting a query without any personal context. Additionally, the framework is efficient for edge deployment, with the client-side model generating only $9.2\%$ of the total tokens. These results demonstrate that \ourmethod{} provides a practical, effective solution for personalized generation with improved privacy.
\end{abstract}




\maketitle

\section{Introduction}

Personalization has become a cornerstone of modern information access systems, including search engines \citep{4604667,10.1145/2488388.2488435, 10.1145/1099554.1099747, 1264820}, recommender systems \citep{Purificato2024UserMA, GUAN201958, lyu-etal-2024-llm}, and question answering (QA) systems \citep{salemi2025pathwaysthoughtsmultidirectionalthinking, salemi2025learningnaturallanguagefeedback} because it enables systems to adapt responses to individual users' preferences, goals, and historical interactions, thereby improving relevance, user satisfaction, and task effectiveness compared to one-size-fits-all models \citep{lamp-qa, salemi2025learningnaturallanguagefeedback}.
Incorporating user-specific context allows these systems to disambiguate underspecified queries, select relevant facts, and control the level of detail and formality, leading to answers that are not only correct but also useful, actionable, and aligned with individual user expectations \citep{lamp-qa}. Prior research has shown that retrieval-augmented generation (RAG) is the current state-of-the-art paradigm for personalizing large language models (LLMs) \citep{lamp, peft-vs-lora, li2025surveypersonalizationragagent, 10.1145/3746252.3760851}, in which a subset of the user profile that is relevant to the query is retrieved and provided as context in order to generate a personalized response.

\begin{figure}[!t]
    \centering
    \includegraphics[width=0.8\linewidth]{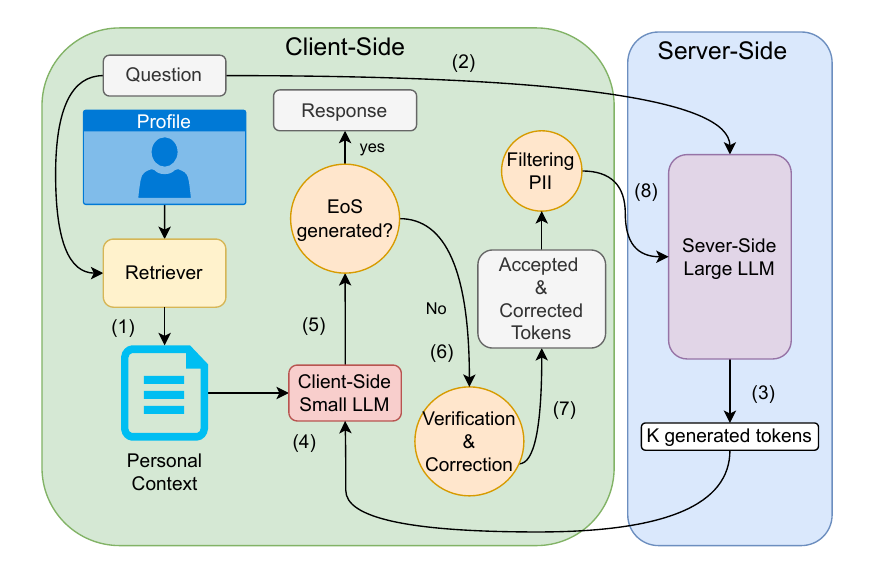}
    \vspace{-0.6cm}
    \caption{An overview of the \ourmethod framework.}
    \vspace{-0.8cm}
    \label{fig:main}
\end{figure}

When retrieval-augmented personalization is applied with a client-side LLM on the user side, it can fully preserve user privacy because the profile never leaves the client's device; however, it is well understood that on-device LLMs must be lightweight due to the strict computational and memory constraints of edge hardware \citep{Xu2024OnDeviceLM, 10.1145/3719664}. These constraints make it impractical to deploy state-of-the-art LLMs locally, as they require levels of computation and memory that are typically unavailable on user devices, which in turn limits performance in applications that benefit from stronger models. Nevertheless, this trade-off is unavoidable in settings such as financial or legal applications, where strict privacy guarantees are mandatory and outweigh the benefits of using more capable server-side models \citep{ojha2025privacy, reynolds2024security, 10.1145/3712001}. However, in many real-world consumer applications, the strict dichotomy between complete local privacy and unrestricted disclosure to a server-side model limits practical deployment. When access to a high-capacity cloud model is necessary for achieving strong performance, the objective shifts from absolute isolation to principled risk reduction. Under this perspective, while some theoretical privacy leakage may be unavoidable, the goal is to minimize the severity of such leakage. This view is consistent with established privacy-preserving frameworks such as $k$-anonymity \citep{10.1142/S0218488502001648, Samarati1998ProtectingPW, 10275805} and differential privacy \citep{10.1145/3640457.3688019, 10275805}, which allow limited, obfuscated, or aggregated information sharing in exchange for high-utility analysis. Following a similar philosophy, this paper targets this class of scenarios and proposes a method for personalizing LLMs while avoiding direct exposure of the user's private personal information to the model provider.

One possible approach is to apply textual \citep{zhang-etal-2025-dyntext, hu-etal-2020-obfuscation, lin-etal-2025-obfuslm} or vector-based \citep{9419784, song2020inversion} obfuscation to sanitize the user profile prior to a transfer to the server-side LLM for personalization. However, this strategy has several limitations. First, such methods are known to be vulnerable to inversion attacks \citep{song2020inversion}, which can partially or fully recover the original profile from its obfuscated form. More critically, if the obfuscated profile remains informative enough to support personalization, then providing it to the provider enables the provider to issue arbitrary queries against it and perform inference about the user without any user interaction, which is effectively equivalent to revealing the original profile and therefore undermines privacy. An alternative is to rely on homomorphic encryption \citep{Rivest1978}, which allows computation to be performed directly on encrypted data without decrypting it; however, existing schemes incur very large computational overheads and are currently impractical for LLM-scale inference, and no major LLM provider supports such functionality in practice. Finally, existing approaches in the literature for improving privacy preservation in LLMs \citep{staab2024beyond, 10.1145/3708821.3733888, 10.1145/3708359.3712156} primarily focus on preventing the model from memorizing or emitting sensitive information from its training data. These methods do not address our concern, where the main privacy risk stems from disclosing the user's personal context to the server as part of the input, rather than from sensitive information appearing in the output.

To address these challenges, this paper introduces the \ourmethod{} framework, shown in Figure~\ref{fig:main}, which enables personalized generation using a server-side LLM without disclosing the user's personal context to the server. The framework decomposes generation into an interactive process between a large server-side model and a small client-side model deployed on the client's device. The server-side model proposes a sequence of $k$ draft tokens as the response for a given user query with no personalization context. The client-side model, which has access to both the query and a set of relevant information retrieved from the user profile, evaluates these draft tokens by comparing their probabilities based on the client-side model probability distribution to the probability of the most likely alternative token. If the relative probability of a draft token falls below a predefined threshold, the token is rejected and replaced with a more appropriate alternative token sampled from the client-side model; otherwise, it is accepted. The client-side model accepts a prefix of verified tokens and, when necessary, modifies the next token before returning control to the server-side model. This iterative speculate-verify-correct procedure continues until an end of response token is generated. Through this process, the client-side model steers the server-side model toward a distribution of outputs that better reflects the user's preferences, while revealing only the minimal information implicit in the accepted or corrected tokens and never exposing the profile itself. If a corrected token contains personally identifiable information (PII) or violates a set of user-specified keywords, it is replaced with a placeholder token before being sent back to the server-side model. This design allows \ourmethod{} to balance strong personalization with practical privacy preservation in settings where users do not fully trust the server but still wish to benefit from large, high-capacity server-side models.

While \ourmethod is applicable to a wide range of personalization tasks, we focus on personalized question answering using the LaMP-QA benchmark \citep{lamp-qa} as our primary testbed because it provides a standardized and realistic evaluation setting where personalization is explicitly required, user profiles are clearly defined, and both utility and privacy leakage can be measured. Our experiments on the recent LaMP-QA benchmark, comprising three diverse personalized question answering datasets, demonstrates that \ourmethod{} consistently outperforms both non-personalized server-side and personalized client-side baselines across all datasets, achieving statistically significant improvements of $7.4\%$ to $9\%$ on average performance depending on the backbone LLM. Notably, \ourmethod{} attains between $90.3\%$ and $95.7\%$ of the performance of a $3-4.6$ times larger ``leaky'' (or insecure) upper-bound setting in which the full personal context is shared with the high-capacity server-side model. Privacy analyses based on linkability and attribute inference attacks show that \ourmethod{} introduces only minimal additional leakage ($1.5\%$ to $3.5\%$) compared to submitting a query alone, offering a substantially safer alternative to standard RAG-based personalization on server-side while achieving higher performance than RAG-based personalization on client-side. Finally, the client-side overhead is low: the client-side model generates only \textasciitilde $9.2\%$ of the final tokens in the response, imposing minimal computational burden on the client side. 

In summary, this paper is the first attempt towards building privacy-preserving retrieval-augmented methods for LLM personalization. Given the dominance of retrieval augmentation solutions in LLM personalization \cite{lamp,longlamp,lamp-qa,salemi2025reasoningenhancedselftraininglongformpersonalized}, we believe that this line of work will have substantial impact on future research in the information retrieval community. To foster future research and encourage development in this area, we open-source the \ourmethod{} implementation.\footnote{Available at: \url{https://github.com/alirezasalemi7/PPP}}

\section{Related Work}

\subsubsection*{\textbf{Personalizing LLMs:}}

Personalization is a central component of modern information access systems, including search engines, recommender systems, and natural language generation models \citep{10.1145/2702123.2702503, 10.1145/1462198.1462203, naumov2019deep, lamp}. In the context of LLMs, \citet{lamp} introduced a retrieval-augmented generation paradigm together with the LaMP benchmark for evaluating short-form personalized text generation. This line of work was later extended to long-form generation in LongLaMP \citep{longlamp} and to personalized question answering in LaMP-QA \citep{lamp-qa}. Personalized conversational and assistant systems have also attracted increasing attention \citep{li2023teach, mysore2023pearl, lu2024corporate, zhang-etal-2024-llm-based}. Existing approaches span a range of strategies, including learning retrieval components from user interaction signals \citep{rspg}, adapting language models using user-specific supervision \citep{jang2023personalized}, and constructing personalized prompts to guide generation \citep{Li_2024}. Parameter-efficient adaptation techniques have also been explored for personalization \citep{tan-etal-2024-personalized}, including their integration with retrieval-augmented architectures \citep{peft-vs-lora}. Beyond direct model adaptation, recent work has shown that incorporating explicit reasoning mechanisms and self-training procedures can further improve personalized text generation \citep{salemi2025reasoningenhancedselftraininglongformpersonalized, salemi2025pathwaysthoughtsmultidirectionalthinking}. Moreover, learning from natural language feedback has been demonstrated to be an effective means of tailoring personalized question answering systems to individual users \citep{salemi2025learningnaturallanguagefeedback}. In contrast to prior studies that focus on better personalizing LLMs, we focus on personalizing an LLM provided by an external provider without disclosing the user's personal context to that provider.

\subsubsection*{\textbf{Privacy in Personalization:}}

Traditionally, privacy-preserving personalization has been studied in contexts where training data from different users are shared to train a model, with methods such as differential privacy or federated learning techniques being proposed \citep{kim-etal-2025-personalized, 10.1145/3733816.3760752, 7023371, 10018494, 10.1145/3640457.3688019}. However, recent research demonstrates that the most effective personalization in text generation is achieved through RAG \citep{lamp, lamp-qa, mysore2023pearl, salemi2025reasoningenhancedselftraininglongformpersonalized}, which introduces privacy risks that extend beyond training. Here, even if a user does not participate in training, sharing personal context with a server-side LLM necessitates disclosing information that can lead to severe privacy leaks. To mitigate these risks, previous work suggests performing personalization entirely on-device so that no information is shared with server-side LLM \citep{peft-vs-lora, 10.1145/3649329.3655665}. However, on-device LLMs must be lightweight due to the strict computational and memory limitations on edge \citep{Xu2024OnDeviceLM, 10.1145/3719664}, often rendering them less capable than high-capacity server-side models. While this trade-off is inevitable for high-stakes applications like banking or financial transactions where privacy constraints are absolute, an intermediate regime exists
---such as managing a personal digital archive or drafting a message---
where privacy is paramount, yet utilizing server LLMs is acceptable as long as the personal context is not shared \citep{LI2025100749}. Common attempts to address this involve textual or vector-based obfuscation \citep{song2020inversion, zeng-etal-2025-mitigating, 10.1145/3336191.3371856, defaveri2024wordsblendingboxesobfuscating}. Nevertheless, these methods are vulnerable to inversion attacks \citep{song2020inversion}, which allow providers to approximately reconstruct the original user profile. More critically, once any form of the profile is disclosed, a provider can probe it with arbitrary queries to infer additional user attributes without the user's involvement. This paper introduces the \ourmethod framework to overcome these challenges by personalizing a cloud LLM without disclosing the personal context in any form as an input to the provider.

\section{The \ourmethod Framework}
\label{sec:method}
\noindent \textbf{Motivation:}
Personalization yields significant gains in question answering, such as improving relevance to user preferences and goals \cite{lamp, longlamp}, increasing answer usefulness and satisfaction \cite{lamp-qa, zhang2025personalization}, adapting responses to the user's background \cite{tseng-etal-2024-two, xu-etal-2025-personalized, 10.1145/3731120.3744584, 10.1145/3626772.3657733}, and better aligning the response with the user's long-term interests \cite{kim2024retrievalenhancedmachinelearningsynthesis, salemi2025reasoningenhancedselftraininglongformpersonalized, salemi-etal-2025-expert, salemi2025learningnaturallanguagefeedback, li2024personalized}. Recent work has shown that the most effective family of methods for personalizing LLMs is based on retrieval augmentation \cite{lamp, rspg, peft-vs-lora, salemi2025pathwaysthoughtsmultidirectionalthinking}, where user-specific context is retrieved from the user profile and provided to the LLM along side the question to guide response generation. However, this approach incurs a privacy cost, as it requires the users to disclose personal context retrieved from their profile to the LLM provider.

\begin{figure}[!t]
    \centering
    \includegraphics[width=\linewidth]{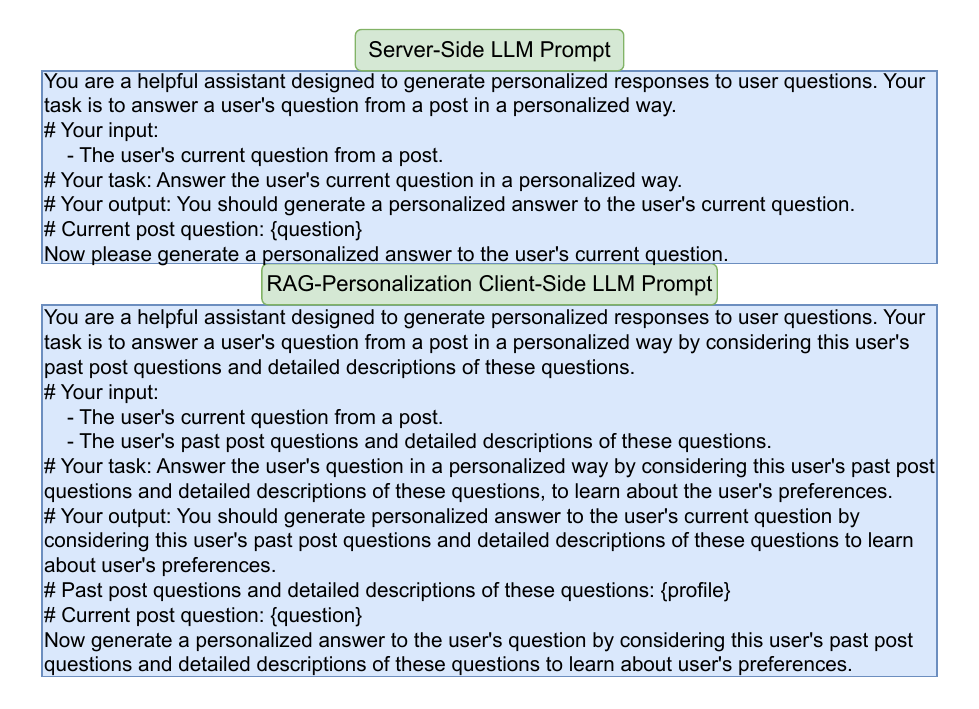}
    \vspace{-1cm}
    \caption{Prompts used with the server-side and user-side LLMs in the \ourmethod framework and the baselines.}
    \label{fig:prompt-ours}
    \vspace{-0.6cm}
\end{figure}

In some scenarios, privacy requirements are strict and disallow sharing of private information with server-side LLMs, often hosted on cloud computing platforms \cite{nguyen2024surveyprivacypreservingmodelexplanations,10478883}. For example, when the personal context contains highly sensitive or irreversible information such as banking credentials, financial records, or proprietary trading strategies. Here, even a small risk of leakage is unacceptable, and only fully local models can be used, despite their limited capabilities \cite{BIBI2025110371, 10.1145/3624010}. However, many real-world applications fall into an intermediate privacy regime, where the strict dichotomy between local isolation and full server-side exposure limits performance. In these scenarios, privacy remains critical, but high model utility motivates a shift from absolute data isolation to principled risk mitigation. This paper adopts this perspective, aiming to substantially reduce privacy leakage while still benefiting from powerful hosted models. Similar to frameworks such as $k$-anonymity \citep{10.1142/S0218488502001648}, which permit controlled data sharing, our approach seeks to balance utility and privacy by enabling personalization without directly disclosing the user's raw personal context to the server-side LLM.

We argue existing solutions suffer from major shortcomings that leave a gap in the literature for building privacy-preserving personalized generation methods. For instance, one might suggest using techniques such as vector-based \cite{song2020inversion} or textual obfuscation \cite{zeng-etal-2025-mitigating,10.1145/3336191.3371856,defaveri2024wordsblendingboxesobfuscating} for sharing an encoded representation or a semantically similar transformation of the user profile with LLM servers to enable personalized generation. However, such approaches are known to be vulnerable to inversion attacks \cite{song2020inversion}, allowing the server to approximately reconstruct the original user profile from its obfuscated form. Even more critically, once any form of the user profile (even obfuscated) is disclosed to the server, it can be reused to generate personalized responses to arbitrary queries generated by the server or adversaries without the user's involvement. This effectively allows the provider to probe the obfuscated profile with its own queries and infer additional information about the user, which violates the intended privacy requirements. An alternative strategy is to use homomorphic encryption \cite{Rivest1978}, which allows computations to be performed directly on encrypted data such that the decrypted result matches the output of the same computation on the plain text. In principle, this would enable the provider to generate a response without ever observing the user's profile or question; however, current homomorphic schemes incur prohibitive computational overheads (often exceeding $1000\times$ slowdowns) and are not supported by existing LLM providers, making it impractical. Moreover, typical approaches \cite{staab2024beyond, 10.1145/3708821.3733888, 10.1145/3708359.3712156} for improving privacy during LLM generation do not apply to our setting, since the primary privacy risk arises from disclosing the user's personal context to the LLM server as part of the input, as opposed to the output.

\medskip
\noindent \textbf{\ourmethod{} Overview:}
To address this problem, we introduce the \ourmethod{} framework (Figure~\ref{fig:main}), which combines a small client-side private model with access to the user profile and a large server-side model provided by a cloud-provider to iteratively and interactively generate a response to the user's query. The client-side model acts as a verifier and modifier for draft token sequences proposed by the server-side model. Specifically, the server-side model with substantially higher cost and generation capability generates a sequence of $k$ draft tokens at a time in response to the user's query without having access to any part of the user's profile. The client-side model then evaluates these tokens conditioned on the user profile, determines how many are acceptable, and, when necessary, replaces the next token with a more appropriate alternative for the user. The server-side model subsequently generates a new draft conditioned on the verified and modified prefix, and this process repeats until an end of response token is selected by the client-side model. This way, the client-side model steers the server-side model which is more capable in text generation toward a distribution of responses that better reflects the user's preferences without revealing the profile itself, by intervening on only a small number of tokens during generation. 

Note that while \ourmethod shares similarities with Speculative Decoding \cite{10.5555/3618408.3619203}, there are several fundamental differences: (1) In \ourmethod, the draft tokens are proposed by the larger server-side model and verified by the smaller client-side model, whereas in speculative decoding the smaller model proposes and the larger model verifies; this reflects our goal of improving personalization and privacy rather than latency. (2) In speculative decoding, both the proposer and verifier are conditioned on the same history, whereas in \ourmethod the two models are conditioned on different inputs: the client-side model additionally observes the user-specific context, while the server-side model does not. (3) Speculative decoding is designed to exactly preserve the output distribution of the large model, whereas \ourmethod intentionally allows controlled deviations from the server-side model's distribution so that the client-side model can steer the generation toward user-specific preferences. (4) In speculative decoding, both the small and large models are deployed on the same side (either locally or server-side), whereas in \ourmethod the server-side large model is hosted by the provider and the client-side model is hosted by the user, which is essential for preserving privacy. The remainder of this section formalizes the problem and describes our method in detail.

\subsection{Problem Formulation}
\label{sec:problem}


We consider a setting in which an LLM provider offers a high-capacity server-side model $M_{\text{server}}$ exclusively via an API, either because the provider does not expose its parameters or because the computational cost of running such a model locally is prohibitive, and a lower-capacity, relatively small client-side model $M_{\text{client}}$ whose parameters are accessible to the user. A user $u$ issues a query $q$ and possesses a private user profile $P_u = \{d_i\}_{i=1}^{|P_u|}$, where each $d_i$ is a personal information item (in the form of text) that the user does not wish to disclose. In this setting, the server-side model is untrusted from a privacy perspective, as it is operated by a third-party provider, while the client-side model is constrained in capacity due to the high computational and memory cost of running large LLMs locally. The objective of this work is to design a method that leverages $P_u$ and $M_{\text{client}}$ to improve the quality of responses generated by $M_{\text{server}}$ for a query $q$, while ensuring that the user's private profile $P_u$ is never directly revealed to the server-side model. The model is evaluated based on the quality of the generated response $\hat{y}$. The evaluation methodology and metrics are presented in Section~\ref{sec:exp-design}.


\begin{algorithm}[!t]
\caption{Implementation of \ourmethod.}\label{alg:our-method}
\begin{algorithmic}[1]
\Require Question $q$, user profile $P_u$, server-side model $M_{\text{S}}$, client-side model $M_{\text{C}}$, retriever $R$, rejection threshold $\tau$, draft length $k$, number of documents $m$
\Ensure Response $\hat{y}$
\State $C_q \gets R(P_u, q, m)$ \Comment{Retrieve personalized context from profile}
\State $\mathcal{E} \gets \text{ExtractPII}(C_q)$ \Comment{Identify PII in the retrieved context}
\State $H_{\text{S}} \gets \text{Template}_{\text{gen}}(q)$ \Comment{Initialize server-side history}
\State $H_{\text{C}} \gets \text{Template}_{\text{rag}}(C_q, q)$ \Comment{Initialize client-side history}
\State $\hat{y} \gets \text{""}$
\While{$\text{EOS} \notin \hat{y}$}
    \State $T_{1:k} \sim M_{\text{S}}(H_{\text{S}}, k)$ \Comment{Server-side model propose $k$ tokens}
    \State $V \gets \emptyset$ \Comment{Accepted token sequence}
    \For{$i \in \{1, ..., k\}$} \Comment{Verification by client-side model}
        \State $P_{T_i} \gets P(T_i \mid H_{\text{C}}, V, T_{1:i-1}; M_{\text{C}})$
        \State $P_{T_i^*}, T^*_i \gets \max_{\omega} P(\omega \mid H_{\text{C}}, V, T_{1:i-1}; M_{\text{C}})$
        \If{$P_{T_i} / P_{T_i^*} < \tau$}
            \State $V \gets V \cup \{T_i^*\}$ \Comment{Client-side model correction}
            \State \textbf{break}
        \Else
            \State $V \gets V \cup \{T_i\}$ \Comment{Accept the draft token}
        \EndIf
    \EndFor
    \State $\hat{y} \gets \hat{y} \text{ ; } V$
    \State $H_{\text{C}} \gets H_{\text{C}} \text{ ; } V$
    \State $H_{\text{S}} \gets \text{FilterPII}(H_{\text{S}} \text{ ; } V, \mathcal{E})$ \Comment{Filter PII in server-side history}
\EndWhile
\State \Return{$\hat{y}$}
\end{algorithmic}
\end{algorithm}

\subsection{Client-Side Modification of Server-Side Generated Speculations}
\label{sec:our-method}

An overview of the \ourmethod framework is shown in Figure~\ref{fig:main}, and the corresponding procedure is given in Algorithm~\ref{alg:our-method}. \ourmethod is interactive and iterative, with computation split between the LLMs on the client and server sides in order to leverage the strong capabilities of the server-side model while preserving user privacy.

When the user issues a query $q$, the first step is performed on the client side. We apply a retrieval model $R$ to select $m$ relevant documents from the user profile $P_u$ to construct the query-specific personal context $C_q = R(P_u, q, m)$, as shown in line~$1$ in Algorithm~\ref{alg:our-method}. This serves two purposes: (1) it filters out irrelevant or distracting information from the profile, retaining only content relevant to the query, and (2) it reduces computational and memory overhead, since user profiles can be large and processing the full profile is costly for resource-constrained user devices with the limited GPU resources and also may exceed model context window limits.

Since the documents in the user profile may contain personally identifiable information (PII) that can cause privacy leakage; we explicitly identify such content to prevent it from being transmitted from the client to the server. As shown in line~$2$ of Algorithm~\ref{alg:our-method}, we detect PII in the retrieved documents. In general, any PII detection approach can be employed, including user-defined keyword lists or existing tools such as \textit{Presidio}\footnote{Available at: \url{https://github.com/microsoft/presidio}}. For simplicity and without loss of generality, we adopt a regex-based method to identify common PII types, including email addresses, phone numbers, IP addresses, URLs, U.S.\ Social Security numbers, credit card numbers, and dates of birth. We apply these patterns (as provided in the accompanying code) to the retrieved documents and extract all matched spans for filtering, denoted as $\mathcal{E} = \text{ExtractPII}(C_q)$.

To generate responses in \ourmethod using $M_{\text{server}}$ and $M_{\text{client}}$, we construct two prompt variants: (1) a server-side prompt (denoted as $H_{\text{S}}$ in Algorithm~\ref{alg:our-method}, line~$3$), instantiated from Figure~\ref{fig:prompt-ours} (top), which contains only the query $q$ and discloses no personal information to the server; and (2) a client-side prompt (denoted as $H_{\text{C}}$ in Algorithm~\ref{alg:our-method}, line~$4$), instantiated from Figure~\ref{fig:prompt-ours} (bottom), which conditions on both the query $q$ and the retrieved personal context $C_q$. Additionally, we initialize the final response as an empty string, denoted as $y = \text{""}$ (line~$5$ of Algorithm~\ref{alg:our-method}). The method then enters an interactive, iterative generation loop that proceeds until the end-of-sequence token (\textit{EOS}) is produced or verified by the client-side model:
\begin{enumerate}[leftmargin=*]
    \item The server-side model $M_{\text{server}}$ generates a sequence of $k$ draft tokens conditioned on the server-side history $H_{\text{S}}$, which comprises the initial server-side prompt and tokens generated and verified in the response up to that point (line 7 in Algorithm~\ref{alg:our-method}).

    \item We denote the sequence of accepted tokens by $V$ (line~$8$ of Algorithm~\ref{alg:our-method}). Then, for each draft token $T_i$ ($1 \leq i \leq k$) generated by $M_{\text{server}}$, we proceed as follows (line~$9$ in Algorithm~\ref{alg:our-method}):
    \begin{enumerate}[leftmargin=*]
        \item We compute the probability that $M_{\text{client}}$ assigns to token $T_i$ conditioned on the prefix $(H_{\text{C}}; T_{1:i-1})$, and denote it by $P_{T_i}$ (line~$10$ of Algorithm~\ref{alg:our-method}). In addition, for the same prefix we obtain the highest-probability token from $M_{\text{client}}$, denoted by $T_i^{*}$ with probability $P_{T_i^{*}}$ (line~$11$ of Algorithm~\ref{alg:our-method}). This step estimates how likely the draft token is under the client-side model $M_{\text{client}}$, which is conditioned on the user's personal context, and compares it to the most probable token by the same model given the personal context.

        \item To determine whether the draft token $T_i$ is acceptable by the client-side model $M_{\text{client}}$, we compare its probability to that of the most likely token $T_i^*$ by $M_{\text{client}}$. If the ratio exceeds a rejection threshold $\tau$, the draft token is accepted; otherwise, it is replaced by $T_i^*$. Formally, if ${P_{T_i}} / {P_{T_i^*}} \geq \tau$, we accept $T_i$ and update the verified sequence as $V = V \cup \{T_i\}$. If ${P_{T_i}}/{P_{T_i^*}} < \tau$, we substitute $T_i$ with $T_i^*$ and update $V = V \cup \{T_i^*\}$. In this case, all remaining draft tokens from the server-side model, $T_{i+1}:T_k$, are discarded and the iteration over draft tokens terminates (lines~$12$--$17$ of Algorithm~\ref{alg:our-method}). This procedure ensures that draft tokens are retained only if they are within a certain margin of error from the most likely token by $M_{\text{client}}$ given the private personal context.
        
    \end{enumerate}

    \item After verification and possible correction, we update the final response by appending the accepted tokens, i.e., $\hat{y} = \hat{y}; V$, and update the client-side history as $H_{\text{C}} = H_{\text{C}}; V$. To update the server-side history, we first append the same tokens and then remove any extracted PII spans $\mathcal{E}$ by replacing them with the \textit{[PII]} marker, yielding $H_{\text{S}} = \text{FilterPII}(H_{\text{S}}; V, \mathcal{E})$. This process is repeated from the first step until the \textit{EOS} token is generated through modification or verified by the client-side model.
\end{enumerate}
At the end of this iterative and interactive procedure between $M_{\text{server}}$ and $M_{\text{client}}$, the response to the user's query is produced as $\hat{y}$. While the verified and corrected tokens are shared back with the server-side model, and therefore may implicitly reveal limited information about the user, this exposure is strictly controlled: PII is filtered (Line~21 in Algorithm~\ref{alg:our-method}), and the private model intervenes only on a small number of tokens. As we show in our privacy analysis in Section~\ref{sec:experiments}, the resulting leakage is minimal---$1.5\%$ in linkability and $3.5\%$ in attribute inference---compared to the non-personalized baseline that only shares the query with the server-side LLM, and is significantly lower than that of RAG personalization baselines that directly expose the profile to the server.

\begin{figure}[!t]
    \centering
    \includegraphics[width=\linewidth]{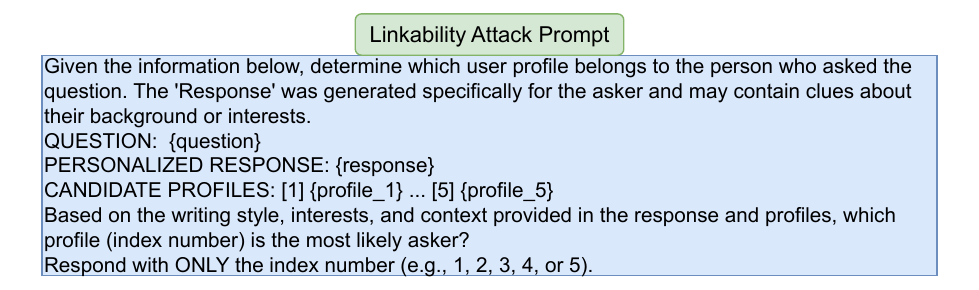}
    \vspace{-0.8cm}
    \caption{Prompt used for the linkability attack with GPT-4.2.}
    \label{fig:linkability-prompt}
    \vspace{-0.4cm}
\end{figure}


\section{Experiments}

\subsection{Experimental Setting}
\label{sec:exp-design}

\begin{figure}[!t]
    \centering
    \includegraphics[width=\linewidth]{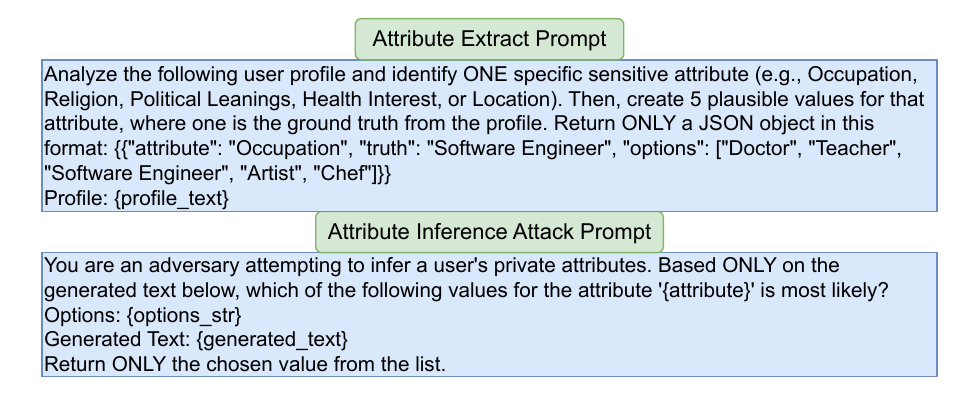}
    \vspace{-0.8cm}
    \caption{Prompts used for the attribute extraction (top) and attribute inference attack (bottom) with GPT-4.2.}
    \label{fig:attribute-inference-prompt}
    \vspace{-0.4cm}
\end{figure}

\begin{table*}[!t]
    \centering
    \caption{Performance of \ourmethod and baselines on the LaMP-QA benchmark. The superscript $^*$ indicates a statistically significant improvement over the best-performing baseline (excluding the setting where personal context is given to the server-side LLM as it serves as the upper bound performance for our method) according to a Student's t-test with $p < 0.05$.}
    \vspace{-0.4cm}
    \adjustbox{max width=0.78\linewidth}{\begin{tabular}{l|ccc|c}
    \toprule
    \multirow{2}{*}{\textbf{Method}} & \textbf{Arts \&} & \textbf{Lifestyle \& Personal} & \textbf{Society \&} & \textbf{Average}  \\
    
    & \textbf{Entertainment} & \textbf{Development} & \textbf{Culture} & \textbf{(macro)} \\
    \midrule
    \multicolumn{5}{c}{Qwen 2.5 Family} \\
    \midrule
    Non-Personalized Server-Side (14B) & 0.3266 & 0.4587 & 0.4870 & 0.4241 \\
    Non-Personalized Speculative Decoding (verifier: 14B; draft: 3B) & 0.3226 & 0.4496 & 0.4770 & 0.4164 \\
    RAG-Personalization Client-Side (3B) & 0.3013 & 0.4278 & 0.4567 & 0.3952 \\
    \midrule
    \midrule
    \ourmethod (Server-Side: 14B; Client-Side: 3B) & \textbf{0.3680$^*$} & \textbf{0.4888$^*$} & \textbf{0.5306$^*$} & \textbf{0.4624$^*$} \\
    \midrule
    \textit{Upper Bound:} RAG-Personalization Server-Side (14B) & 0.3920 & 0.5107 & 0.5743 & 0.4923 \\
    \midrule
    \multicolumn{5}{c}{Gemma 3 Family} \\
    \midrule
    Non-Personalized Server-Side (12B) & 0.4132 & 0.5954 & 0.6299 & 0.5461 \\
    Non-Personalized Speculative Decoding (verifier: 12B; draft: 4B) & 0.3912 & 0.5799 & 0.5923 & 0.5211 \\
    RAG-Personalization Client-Side (4B) & 0.4258 & 0.5765 & 0.6013 & 0.5345 \\
    \midrule
    \midrule
    \ourmethod (Server-Side: 12B; Client-Side: 4B) & \textbf{0.4811$^*$} & \textbf{0.6199} & \textbf{0.6592$^*$} & \textbf{0.5867$^*$} \\
    \midrule
    \textit{Upper Bound:} RAG-Personalization Server-Side (12B) & 0.5327 & 0.6722 & 0.7141 & 0.6396 \\
    \bottomrule
    \end{tabular}}
    \label{tab:main-results}
    \vspace{-0.4cm}
\end{table*}

\subsubsection*{\textbf{Datasets and Evaluation Methodology:}}

We conduct experiments on the LaMP-QA benchmark \citep{lamp-qa}, a recent and the only publicly available benchmark for personalized question answering. LaMP-QA covers three datasets: (1) Art \& Entertainment (767 questions), (2) Lifestyle \& Personal Development (989 questions), and (3) Society \& Culture (1074 questions).\footnote{Other datasets for personalized text generation, such as LaMP \citep{lamp} and LongLaMP \citep{longlamp}, are not suitable for our setting: First, these datasets primarily focus on mimicking user writing style, which is not well suited for studying privacy preservation, since privacy leakage is difficult to quantify from stylistic features alone. Second, unlike the LaMP-QA benchmark, which provides explicit annotations of user expectations, they include only a single reference response. This limits their ability to capture the diversity of valid personalized outputs and can lead to misalignment with the user's real expectations about the response, as clearly noted by \citet{lamp-qa}.} Each LaMP-QA example consists of a user query, the user's history (serving as a profile) consisting of the past questions and their detailed information need the user has asked, a narrative describing the user's intent and perspective, and a set of personalized rubrics specifying the aspects an ideal response should address. For evaluation, we follow the protocol of \citet{lamp-qa} and use Qwen~2.5~(32B) \cite{qwen2025qwen25technicalreport} as the judge model. For each query, the judge evaluates whether the generated response satisfies each personalized aspect, assigning scores in the range $[0,2]$. These scores are normalized to $[0,1]$ by being divided by $2$, and the final response score is computed as the mean normalized score across all aspects for the user's question. We refer the reader to \citet{lamp-qa} for further details.

\subsubsection*{\textbf{Inference Setting:}}

For $M_{\text{server}}$, we use instruction-tuned Qwen~2.5 with 14B parameters\footnote{Available at: \url{https://hf.co/Qwen/Qwen2.5-14B-Instruct}} \cite{qwen2025qwen25technicalreport} and instruction-tuned Gemma~3 with 12B parameters\footnote{Available at: \url{https://hf.co/google/gemma-3-12b-it}} \cite{gemmateam2025gemma3technicalreport}, unless otherwise specified. Although our method is applicable to proprietary models such as GPT and Gemini, their current APIs do not support sampling $k$ tokens and resuming generation from an intermediate state; instead, generation must be restarted with a new user message, which we found to introduce inconsistencies. Therefore, we do not include these models in our experiments. As $M_{\text{client}}$, we use smaller models from the same families, namely instruction-tuned Qwen~2.5 with 3B parameters\footnote{Available at: \url{https://hf.co/Qwen/Qwen2.5-3B-Instruct}} and instruction-tuned Gemma~3 with 4B parameters.\footnote{Available at: \url{https://hf.co/google/gemma-3-4b-it}} For \ourmethod, we set the number of draft tokens generated from $M_{\text{server}}$ to $k=10$, unless otherwise specified, and set the rejection threshold to $\tau=0.05$. We use nucleus sampling \cite{Holtzman2020The} with temperature $1.0$ for all models. The maximum context length (inputs and outputs) is set to 2048 for $M_{\text{server}}$ and 8192 for $M_{\text{client}}$. For retrieval over profiles, we use Contriever \cite{izacard2022unsupervised} to retrieve $m=10$ documents. We use vLLM \cite{vllm} for inference. All experiments are conducted on two NVIDIA A100 GPUs with 80GB of memory each and 256GB of system RAM.

\subsubsection*{\textbf{Baselines:}}

To evaluate the efficacy and privacy of our proposed method, we compare it against the following architectural baselines:

\begin{itemize}[leftmargin=*]
    \item \textbf{Non-Personalized (Server-Side Only):} A setting in which the most capable server-side model $M_{\text{server}}$ directly answers queries without access to the user's private profile. This corresponds to the current industry standard for privacy-sensitive, API-based interactions. We use instruction-tuned Qwen~2.5-14B and Gemma~3-12B in this setting, similar to the backbone LLMs in our method.

    \item \textbf{Non-Personalized Speculative Decoding:} We apply speculative decoding \cite{10.5555/3618408.3619203} with the same hyperparameters and backbone LLMs as \ourmethod, due to its structural similarity.\footnote{Although speculative decoding and \ourmethod differ substantially, as discussed in details in Section~\ref{sec:method} of our paper, we include it as a baseline because both involve a small model and a larger model that generate responses collaboratively.} This is similar to the non-personalized baseline, except that the client-side model generates draft tokens without conditioning on personal information and sends them to the server-side model for verification. For details on speculative decoding, we refer the reader to \cite{10.5555/3618408.3619203}.

    \item \textbf{Client-Side RAG-Personalization:} A fully private baseline based on \citet{lamp}, in which a small model $M_{\text{client}}$ (Qwen~2.5--3B and Gemma~3--4B, same backbone as our method for a fair comparison) performs both retrieval and generation on client-side. This baseline serves as a fully privacy-preserving point of comparison for evaluating the effectiveness of our method.

\end{itemize}
We also use the following upper bound as a term of comparison:
\begin{itemize}[leftmargin=*]
    \item \textbf{Server-Side RAG-Personalization (Upper Bound):} To assess the personalization quality, we include a ``leaky'' baseline where the full private context is transmitted to the server-side model. This serves as the utility upper bound, allowing us to measure the ``personalization gap'' introduced by our privacy constraints.
\end{itemize}
Note that methods like $k$-anonymity \citep{10.1142/S0218488502001648, Samarati1998ProtectingPW, 10275805}, federated learning \citep{lee2023fedlp}, and differential privacy \citep{10.1145/3640457.3688019, 10275805} are primarily used in LLM training and aim to prevent training data from being revealed at inference. Conversely, our focus is on inference without training, where personal information is explicitly provided during inference. Therefore, they are not considered baselines in our evaluation.

\begin{figure*}[!t]
    \centering
    \includegraphics[width=0.8\linewidth]{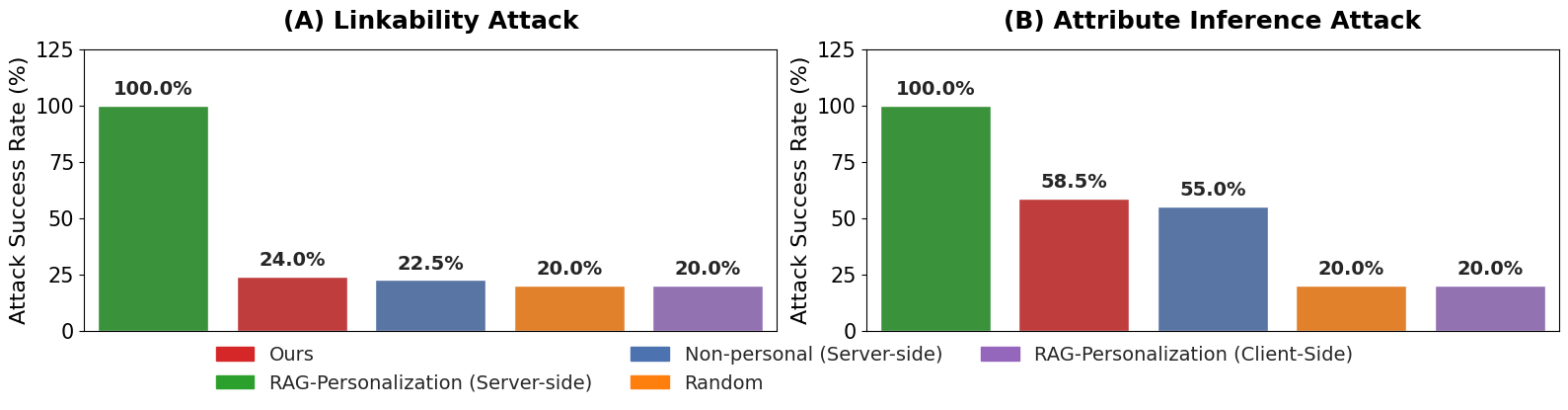}
    \vspace{-0.5cm}
    \caption{Effectiveness of privacy attacks on different methods (lower is better). (A) Linkability attack, in which the cloud LLM provider uses information collected during response generation to identify the true user profile among five candidates. (B) Attribute inference attack, in which the cloud LLM provider uses information collected during response generation to infer user attributes from five possible options per attribute. Experiments are conducted using the Qwen-2.5 family as the backbone models (3B as the client-side and 14B as the server-side LLM) for the evaluated methods and GPT-4.2 as the attacker model.}
    \label{fig:mia}
    \vspace{-0.4cm}
\end{figure*}

\subsection{Main Findings}
\label{sec:experiments}

\subsubsection*{\textbf{On Personalization Performance:}}

The results on the LaMP-QA benchmark are presented in Table~\ref{tab:main-results}. We observe that regardless of the backbone LLM, the small client-side personalized model on the user side underperforms the non-personal server-side model provided by the cloud-provider on average as well as nearly all individual settings, which motivates leveraging the non-personal server-side model rather than relying solely on a personalized client-side model. Furthermore, Table~\ref{tab:main-results} shows that \ourmethod outperforms all baselines across all datasets and on average with statistical significance, demonstrating the effectiveness of our approach for improving personalization without sharing personal information with the server. \ourmethod achieves between 90.3\% and 95.7\% of the performance of a setting in which personal context is directly shared with the large server-side model, which is the upper bound personalization performance on this task. This indicates that strong personalization can be obtained while protecting user profile. We attribute this behavior to the complementary strengths of the two models: the server-side model is more capable of generating high-quality and fluent general responses, while the client-side model, by conditioning on personal information, is better suited to guiding the response toward the user's preferences. Their combination allows the system to generate accurate response while maintaining alignment with the user's personal context, resulting in improved overall performance.

\subsubsection*{\textbf{On Privacy Analysis:}}

To study the privacy properties of \ourmethod for personalization, we follow \citet{staab2024beyond} and conduct two sets of experiments. In both, we evaluate the extent to which the provider of the server-side model can infer or recover information about the user based on the information it observes during the process:

\textbf{(A) Linkability Attack:} In this experiment, we evaluate how well an attacker can identify the user profile associated with a given generated response from a set of candidate profiles. We sample 200 examples from the LaMP-QA benchmark. For each example, we retrieve 20 candidate user profiles based on textual similarity to the true profile using BM25 \cite{Robertson1994OkapiAT}, and then randomly select four of them as negative candidates.\footnote{We observed that selecting random profiles without conditioning on similarity allows the attacker to identify the correct profile with nearly 100\% accuracy based on the query alone, which is unrealistic. Conditioning the candidates on similarity yields a more challenging and realistic attack that tests whether the attacker can distinguish the true user among plausible alternatives.} The resulting set consists of one true profile and four negatives. Given the query, the generated response, and the five candidate profiles, we prompt a large and capable off-the-shelf LLM, GPT-4.2 \cite{openai2024gpt4technicalreport} (prompt in Figure~\ref{fig:linkability-prompt}) to predict which profile the response corresponds to.\footnote{\label{fn:setting}For fully privacy-preserving baselines (i.e., the client-side private model), no information is provided to the attacker, and predictions are therefore random.}This allows us to quantify how effectively an attacker can infer the identity of the user from the response, and thus to measure the degree of personal information leakage (the lower the accuracy, the less leakage). The results of this experiment are shown in Figure~\ref{fig:mia}(A). As expected, the highest level of leakage occurs when the full personal context is directly provided to the server-side model, since the provider observes the entire user profile. In contrast, a random baseline and the fully private setting (in which only the client-side model is used and no information is shared with the provider) yield the lowest identification accuracy, at 20\%. Submitting only the query to the server-side model yields an identification accuracy of 22.5\%, while \ourmethod yields 24\%. Thus, \ourmethod increases the attacker's accuracy by only 1.5 percentage points relative to the query-only setting, indicating that it reveals very little additional information. Moreover, its leakage is only 4 percentage points higher than the fully private or random baseline. These results indicate that \ourmethod achieves statistically significant improvements in personalization while incurring only minimal additional privacy leakage.

\begin{figure*}[!t]
    \centering
    \includegraphics[width=0.85\textwidth]{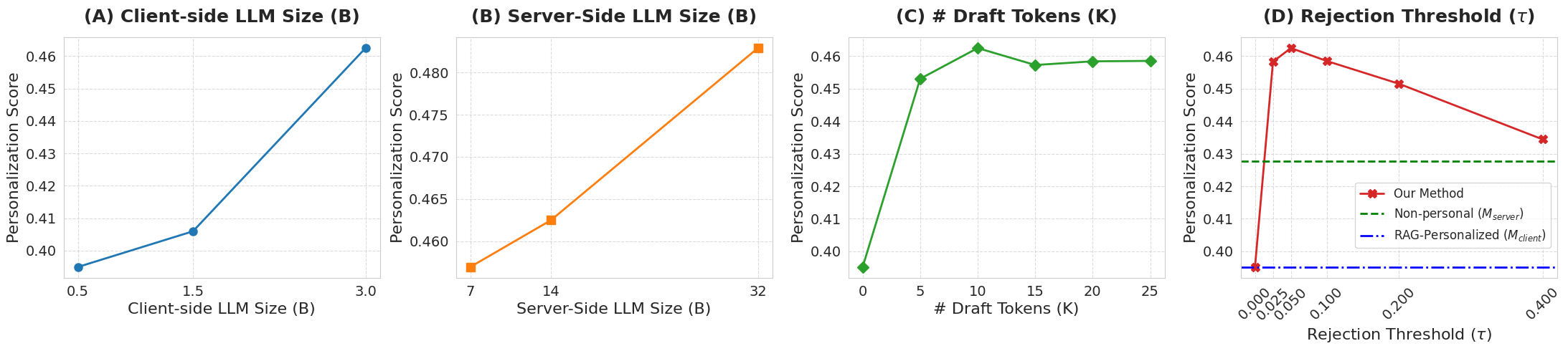}
    \vspace{-0.5cm}
    \caption{Ablation study of \ourmethod with Qwen~2.5 as the backbone LLM: (A) Effect of client-side LLM size (\#parameters in billion) on performance. (B) Effect of server-side LLM size (\#parameters in billion) on performance. (C) Effect of the number of sampled tokens ($k$) from the server-side LLM on performance. (D) Effect of the rejection threshold ($\tau$) on performance.}
    \label{fig:combined}
    \vspace{-0.4cm}
\end{figure*}

\begin{figure}[!t]
    \centering
    \includegraphics[width=0.85\linewidth]{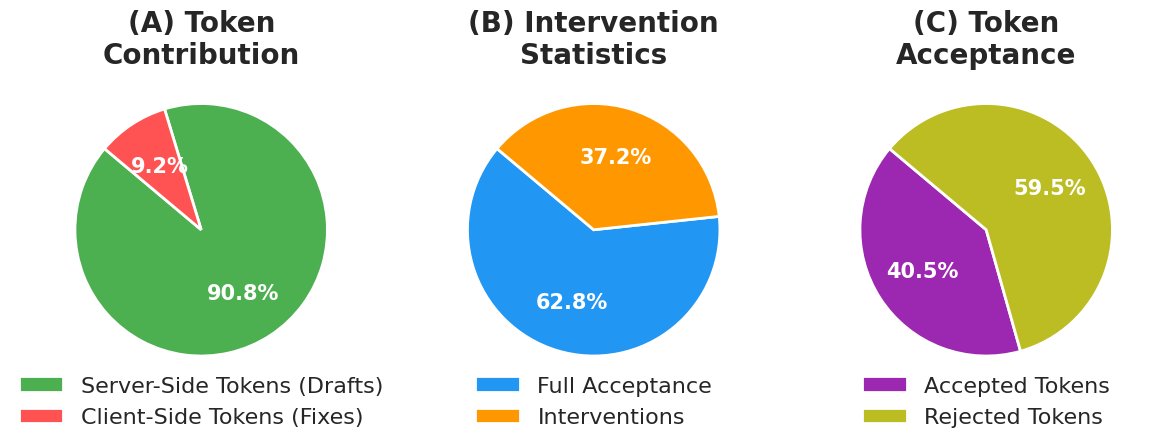}
    \vspace{-0.5cm}
    \caption{Token usage and intervention in \ourmethod. (A) Proportion of tokens generated by the server- and client-side models. (B) Frequency of client-side model interventions. (C) Proportion of server-side tokens accepted when intervention occurs.}
    \label{fig:stats}
    \vspace{-0.4cm}
\end{figure}

\textbf{(B) Attribute Inference Attack:} In this experiment, we evaluate how well an attacker (i.e., the server-side model provider) can infer user attributes based on the information it observes during response generation. We sample 200 examples from the LaMP-QA benchmark. For each query-profile pair, we use GPT-4.2 with the prompt shown in Figure~\ref{fig:attribute-inference-prompt} (top) to extract a single attribute from the personal context, chosen from \{Occupation, Religion, Interests, Location, Nationality, Political Leanings\}, along with its corresponding value and a set of plausible alternative values. These candidates are validated via human judgment to ensure correctness. We then provide GPT-4.2 with the query and the generated response (i.e., the information visible to the server-side model during generation), using the prompt shown in Figure~\ref{fig:attribute-inference-prompt} (bottom),  to predict the correct attribute value.\footnote{Similar to the setting described in Footnote~\ref{fn:setting}.} This measures the extent to which personal information embedded in the generated responses (i.e., information visible to the cloud LLM provider) enables inference of user attributes, and thus quantifies privacy leakage. The results of this experiment are shown in Figure~\ref{fig:mia}(B). As expected, the highest inference accuracy occurs when the full personal context is provided to the server-side model, since the provider observes the entire user profile and can predict attributes with 100\% accuracy. In contrast, the random baseline and the fully private setting yield the lowest accuracy, as no information is shared with the provider and the attacker's predictions are effectively random. For \ourmethod, the attacker achieves an accuracy of 58.5\%, which is only 3.5 percentage points higher than the non-personal baseline that provides only the query to the server-side model. This indicates that a substantial portion of attribute inference is possible from the query alone (55\%), and that \ourmethod introduces only a small additional leakage relative to this baseline. At the same time, this leakage is far lower than that incurred by directly sharing the full personal context with the server-side model. These results suggest that \ourmethod achieves a favorable trade-off between personalization and privacy: it yields statistically significant improvements in personalization while introducing only minimal additional attribute leakage beyond what is already revealed by the query itself. However, they also highlight that even submitting a query to a server-side model can reveal non-trivial information about the user,\footnote{One may ask whether leakage in the non-personal baseline is primarily due to lexical cues in the user's query. To examine this, we conduct an experiment in which we first paraphrase each query using GPT-4.2 and then submit it to the server-side model. We observe similar levels of attribute inference in this setting, indicating that the leakage is driven by the semantic content of the query rather than by specific word choices.} and thus users concerned about these attributes should avoid server-side LLMs.

\subsubsection*{\textbf{On the Size of Client-Side LLM in \ourmethod:}}

To study how the size of the client-side model affects personalization performance, we use instruction-tuned Qwen~2.5 models with 0.5B, 1.5B, and 3B parameters as $M_{\text{client}}$, and instruction-tuned Qwen~2.5--14B as $M_{\text{server}}$ in \ourmethod. The results are shown in Figure~\ref{fig:combined}~(A), which reports average performance across all datasets in LaMP-QA. The results indicate that increasing the size of the client-side model improves performance. However, this improvement comes at a cost: the average runtime of \ourmethod per query is $14.2$ seconds for the $0.5$B model, $19.2$ seconds for the $1.5$B model, and $23.6$ seconds for the $3$B model, corresponding to increased latency. We note that this computational cost is split (as detailed later) between the client-side and the server-side.

\subsubsection*{\textbf{On the Size of Server-Side LLM in \ourmethod:}}

To study how the size of the server-side model affects personalization performance, we fix $M_{\text{client}}$ to instruction-tuned Qwen~2.5--3B and vary $M_{\text{server}}$ among instruction-tuned Qwen~2.5 models with 7B, 14B, and 32B parameters. The results are shown in Figure~\ref{fig:combined}~(B), which reports average performance across all datasets in LaMP-QA. The results show that increasing the size of the server-side model improves performance. This improvement comes at the cost of increased latency: the average runtime per query of \ourmethod is 16.3 seconds for the 7B model, 23.6 seconds for the 14B model, and 25.6 seconds for the 32B, which are divided between the client-side model and the server-side model.

\begin{figure}[!t]
    \centering
    \includegraphics[width=0.85\linewidth]{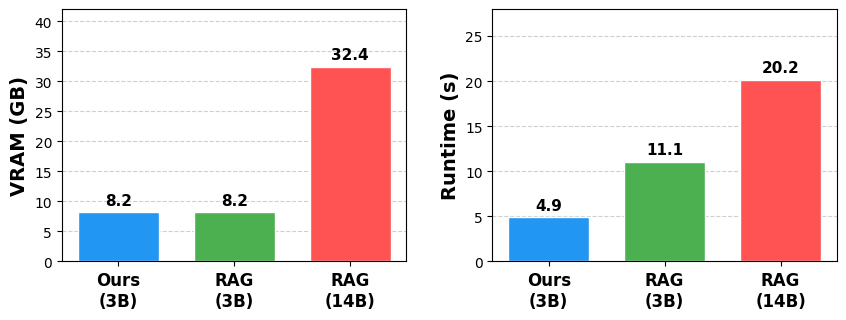}
    \vspace{-0.5cm}
    \caption{User-side VRAM (GB) and runtime (seconds) comparison of \ourmethod and RAG-Personalization with different Qwen 2.5 LLM sizes (3B and 14B) as the backbone LLM.}
    \label{fig:vram-runtime-stats}
    \vspace{-0.5cm}
\end{figure}

\begin{figure*}[!t]
    \centering
    \includegraphics[width=0.95\textwidth]{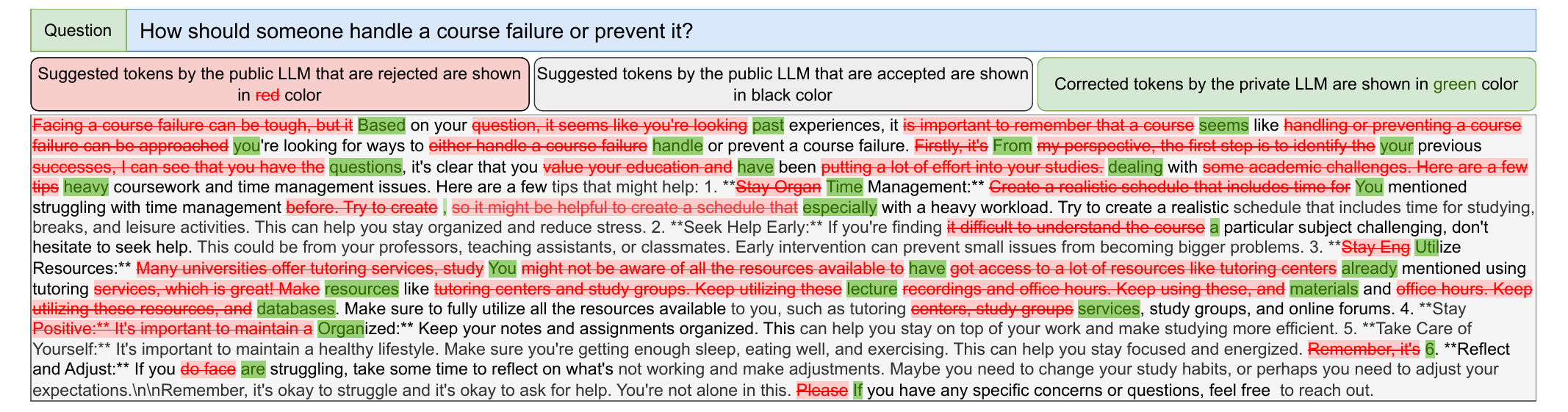}
    \vspace{-0.5cm}
    \caption{Case study of a response generated by \ourmethod with Qwen 2.5 family of models as the backbone LLM.}
    \label{fig:case-study}
    \vspace{-0.4cm}
\end{figure*}

\subsubsection*{\textbf{On Hyperparameter Sensitivity:}}

To study the effect of hyperparameters of \ourmethod on personalization performance, we conduct two experiments. First, we vary the number of draft tokens $k$ sampled from $M_{\text{server}}$ at each step, with $k \in \{0, 5, 10, 15, 20, 25\}$, where $k=0$ corresponds to generating all tokens from the client-side model. The average performance on all datasets in the LaMP-QA benchmark are shown in Figure~\ref{fig:combined}~(C) using instruction-tuned Qwen~2.5--3B as $M_{\text{client}}$, instruction-tuned Qwen~2.5--14B as $M_{\text{server}}$, and a rejection threshold of $\tau=0.05$. The results show that increasing $k$ improves performance up to a point, after which performance saturates and slightly declines; this decline is not statistically significant under a Student's $t$-test with $p<0.05$. This suggests that an intermediate value of $k$ yields optimal performance. In the second experiment, we fix $M_{\text{client}}$ to instruction-tuned Qwen~2.5--3B, $M_{\text{server}}$ to instruction-tuned Qwen~2.5--14B, and $k=10$, and vary the rejection threshold $\tau \in \{0, 0.025, 0.05, 0.1, 0.2, 0.4\}$, where $\tau=0$ again corresponds to using only client-side tokens. The average performance on all datasets in the LaMP-QA benchmark are shown in Figure~\ref{fig:combined}~(D). When $\tau$ is small, performance is close to that of the client-side model, reflecting the strict acceptance criterion. Increasing $\tau$ improves performance up to $\tau=0.05$, after which performance degrades. For larger values of $\tau$, the system accepts nearly all tokens from the server-side model, which has no access to personal information, and performance converges toward that of the non-personalized baseline. These results indicate that an intermediate value of $\tau$ is necessary to balance the contributions of the client- and server-side models and achieve optimal personalization performance.

\subsubsection*{\textbf{On Client-Side Computation Costs:}}

An important consideration for on-device systems such as the client-side model $M_{\text{client}}$ is that they should impose minimal computational overhead to the clients \cite{10.1145/3711875.3729132, xu-etal-2025-personalized}. To analyze this, we report statistics using the Qwen~2.5 backbone LLM on 200 sampled examples from the LaMP-QA benchmark. As shown in Figure~\ref{fig:stats}(A), only 9.2\% of the tokens in the final responses are generated by the client-side model. On average, responses contain 253.3 tokens, of which only 23.3 are produced by $M_{\text{client}}$, with the remainder coming from $M_{\text{server}}$. In terms of runtime, the average latency per query is 23.6 seconds, of which only 4.9 seconds are spent on the client-side model, while the rest is attributable to the server-side model. This indicates that the majority of the computational burden resides on the provider side. Furthermore, as shown in Figure~\ref{fig:stats}(B), in 62.8\% of cases the client-side model accepts all tokens proposed by the server-side model and does not intervene, while interventions occur in only 37.2\% of cases. This suggests that $M_{\text{client}}$ primarily acts as a guiding or filtering mechanism rather than as the main generator. Finally, Figure~\ref{fig:stats}(C) shows that when interventions do occur, the client-side model accepts on average 40.5\% of the proposed tokens, corresponding to approximately 4.5 tokens when $k=10$. Thus, even in cases where intervention is required, a substantial fraction of the server-side model's suggested tokens is still retained by the client-side model. Overall, these results indicate that \ourmethod places minimal computational load on the user side while leveraging the server-side for the bulk of generation. To further compare against the strongest fully privacy-preserving baseline, namely RAG-Personalization with a client-side model, we measure user-side runtime and VRAM consumption for both methods (Figure~\ref{fig:vram-runtime-stats}). The results indicate that \ourmethod with a 3B client-side model significantly outperforms RAG-Personalization with a 3B model while using the same amount of VRAM and less user-side runtime, since most generation is delegated to the server-side model. Achieving comparable performance with RAG-Personalization requires increasing the client-side model size to 14B (matching the size of $M_{\text{server}}$ in \ourmethod), which increases user-side VRAM usage to 32.4\,GB and runtime to 20.2 seconds per query, imposing substantial overhead on the user compared to \ourmethod. These results demonstrate that \ourmethod offers a favorable trade-off among personalization quality, privacy preservation, and user-side computational efficiency.

\subsection{Case Study}

As a case study of \ourmethod, we analyze an example from LaMP-QA in which a user seeks advice on coping with academic failure (Figure~\ref{fig:case-study}). This example illustrates how the client-side model steers a general response toward a user-specific one. When the server-side model proposes a generic opening, the client-side model replaces it with a more empathetic acknowledgment that reflects the user's history (e.g., prior successes and heavy coursework), thereby grounding the response in the user's context. As a result, recommendations such as focusing on time management are framed as direct responses to the user's documented challenges rather than as generic advice. Importantly, the client-side model's edits serve two roles: they introduce relevant personal context and shift the response distribution toward more appropriate outputs for the user. Not all edits correspond to the insertion of personal facts; some refine tone or clarity, for example by replacing vague suggestions such as ``staying positive'' with more concrete and actionable guidance such as ``staying organized,'' informed by the user's history. These refinements are achieved efficiently: the client-side model generates only a small fraction of the final tokens and functions primarily as a selective editor rather than as the main generator. This case study demonstrates that a small client-side model can substantially improve the usefulness and appropriateness of a large server-side model's outputs by providing a context-aware final layer of control.

\section{Conclusions}

This paper studies the problem of personalizing LLMs in a setting where users are unwilling to disclose their personal context to a server-side model due to privacy and trust concerns. We propose \ourmethod{}, an interactive framework that combines a relatively small client-side model with a large server-side model to generate personalized responses without sharing user profiles. The server-side model proposes fluent draft tokens, which the client-side model verifies and selectively modifies them based on the user profile, steering generation toward user-appropriate outputs while preserving privacy. Extensive Experiments show that \ourmethod{} consistently outperforms non-personalized and fully client-side personalized baselines across all personalized QA datasets in the LaMP-QA benchmark, and achieves a large fraction of the performance of a setting in which the profile is directly shared with the server-side model. Privacy analyses based on linkability and attribute inference attacks indicate that \ourmethod{} introduces only minimal additional leakage beyond submitting the query alone to the server-side model, and substantially less leakage than approaches that expose the full profile. We further show that the user-side overhead is small: only a minor fraction of tokens are generated or replaced by the client-side model, with most computation remaining on the server-side. Overall, \ourmethod{} offers a practical trade-off between utility, privacy, and efficiency, enabling users to benefit from powerful cloud-hosted LLMs while retaining control over their personal data, and motivating further work on interactive, privacy-aware architectures for personalized QA.

\section*{Acknowledgment}

This work was supported in part by the Center for Intelligent Information Retrieval, in part by NSF grant \#2143434, in part by the Office of Naval Research contract \#N000142412612, and with support from by Google.org. Any opinions, findings and conclusions or recommendations expressed in this material are those of the authors and do not necessarily reflect those of the sponsor.

\bibliographystyle{ACM-Reference-Format}
\bibliography{XX-references}

\end{document}